\pdfoutput=1

\documentclass[11pt]{article}

\usepackage{EMNLP2023}

\usepackage{times}
\usepackage{latexsym}

\usepackage[T1]{fontenc}

\usepackage[utf8]{inputenc}

\usepackage{microtype}

\usepackage{inconsolata}

\usepackage{array}
\usepackage{CJKutf8}
\usepackage{booktabs}
\usepackage{graphicx}
\usepackage{multirow}
\usepackage{xspace}
\usepackage{amsfonts} 
\usepackage{soul}
\usepackage{makecell}
\newcommand{\silver}[0]{\textsc{AuthorAnchors}\xspace}
\newcommand{\gold}[0]{\textsc{ReaderAnchors}\xspace}
\definecolor{examplepurple}{HTML}{9900ff}
\definecolor{anchoryellow}{HTML}{fff2cc}
\DeclareRobustCommand{\hlanchor}[1]{{\sethlcolor{anchoryellow}\hl{#1}}}

\newcommand{\emldisplay}[2]{\texttt{\href{mailto:#1}{#2}}}
\newcommand{\eml}[1]{\emldisplay{#1}{#1}}
\newcommand{\finalversion}[1]{\unskip}
\let\svthefootnote\thefootnote
\newcommand\blankfootnote[1]{%
  \let\thefootnote\relax\footnotetext{#1}%
  \let\thefootnote\svthefootnote%
}

\newif\ifcomments

\ifcomments
    \providecommand{\nfl}[1]{{\protect\color{Green}{[NFL: #1]}}}
    \providecommand{\kl}[1]{{\protect\color{ProcessBlue}{[KL: #1]}}}
    \providecommand{\kt}[1]{{\protect\color{Red}{[KT: #1]}}}
\else
    \providecommand{\nfl}[1]{}
    \providecommand{\kl}[1]{}
    \providecommand{\kt}[1]{}
\fi

\title{Anchor Prediction: Automatic Refinement of Internet Links}

\author{Nelson F. Liu \\
  Stanford University \\
  \eml{nfliu@cs.stanford.edu} \\\And
  Kenton Lee \and Kristina Toutanova \\
  Google DeepMind \\
  \{\emldisplay{kentonl@google.com}{kentonl},\emldisplay{kristout@google.com}{kristout}\}\texttt{@google.com} \\}

\begin{document}
\maketitle
\blankfootnote{\llap{\textsuperscript{*}}Work completed as a student researcher at Google.}
\begin{abstract}
Internet links enable users to deepen their understanding of a topic by providing convenient access to related information.
However, the majority of links are \emph{unanchored}---they link to a target webpage as a whole, and readers may expend considerable effort localizing the specific parts of the target webpage that enrich their understanding of the link's source context.
To help readers effectively find information in linked webpages, we introduce the task of \emph{anchor prediction}, where the goal is to identify the specific part of the linked target webpage that is most related to the source linking context.
We release the \silver{} dataset, a collection of 34K naturally-occurring anchored links, which reflect relevance judgments by the authors of the source article. To model reader relevance judgments, we annotate and release \gold{}, an evaluation set of anchors that readers find useful. Our analysis shows that effective anchor prediction often requires jointly reasoning over lengthy source and target webpages to determine their implicit relations and identify parts of the target webpage that are related but not redundant.
We benchmark a performant T5-based ranking approach to establish baseline performance on the task, finding ample room for improvement.
\end{abstract}

\section{Introduction}

Links are an indispensable tool for finding information on the web, enabling users to deepen their understanding of a topic by providing convenient access to related webpages.
However, the majority of links are \emph{unanchored}---they simply point to another webpage in its entirety.
Unanchored links to lengthy target webpages leave readers with the time-consuming and error-prone task of scrolling through to identify the information relevant to the source linking context.

\begin{figure}
  \centering
  \includegraphics[width=0.9\columnwidth]{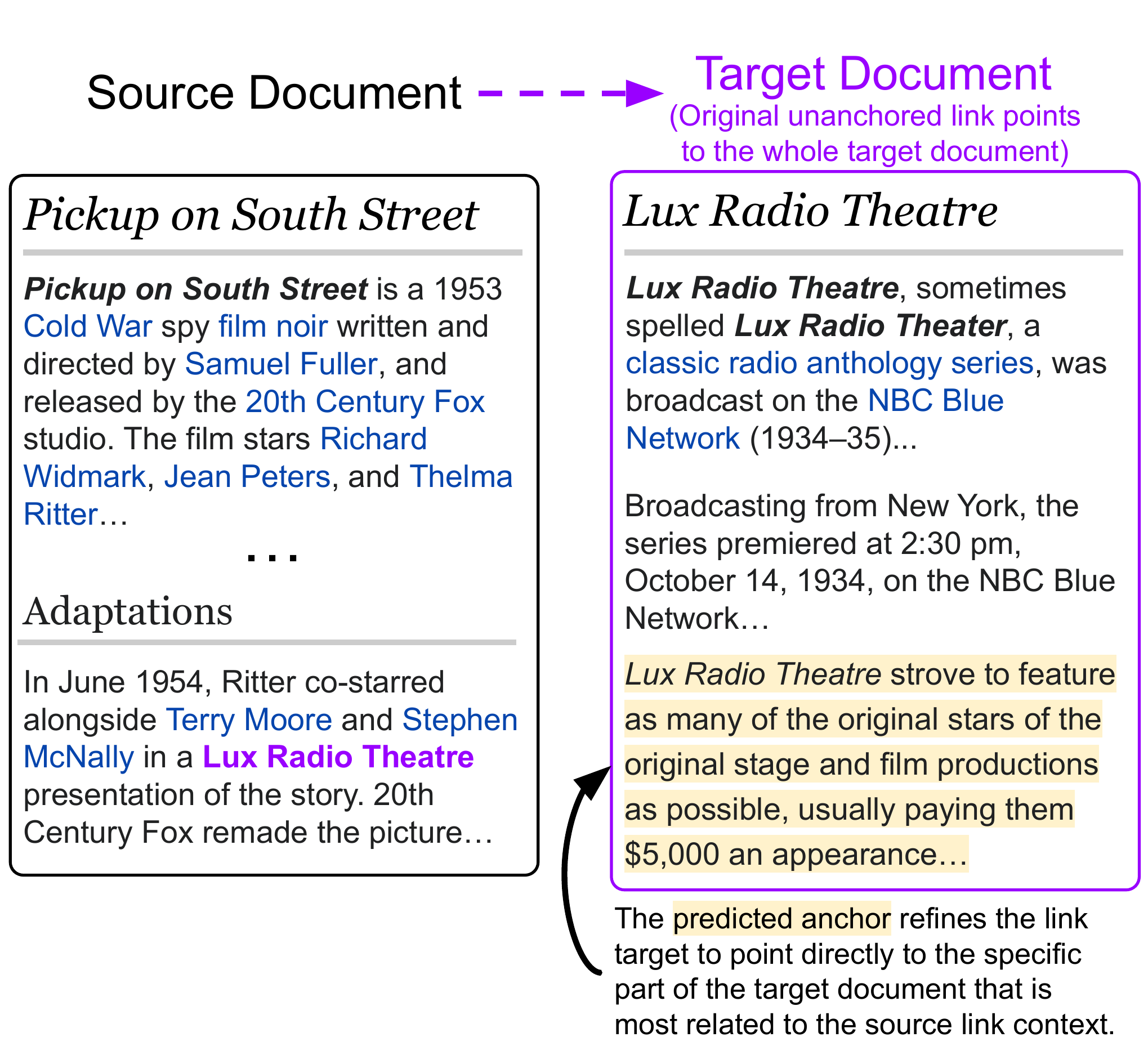}
  \caption{The majority of links are \emph{unanchored}---they point to a target webpage in its entirety. In the task of \emph{anchor prediction}, (i)~a {\color{examplepurple} link}, (ii)~its context in the source webpage, and (iii)~the target webpage are provided as input, and the goal is to find the specific part of the target webpage (the \hlanchor{anchor}) that is most related to the link's source context.}
  \label{fig:example}
\end{figure}

To help readers effectively find information on the web, we introduce the task of \emph{anchor prediction}, where the goal is to refine the target of an unanchored link by identifying the specific part of the target webpage (the \emph{anchor}) that is most related to the source linking context. An example is provided in Figure~\ref{fig:example}.
Given a link and a set of candidate refined anchors, a model must select the candidate that is most relevant to the link within its particular source context. 

Beyond its utility as a standalone application, effective anchor prediction poses unique language understanding challenges---models must jointly reason over multiple parts of lengthy source and target webpages in order to (i)~understand the relation between the source webpage, the link text, and the target webpage, (ii)~determine what evidence is relevant and which can be ignored, and (iii)~identify the most relevant candidate anchor. For example, refining the link in Figure~\ref{fig:example} requires reading multiple parts of the source webpage to understand that Ritter starred in both the original film and the \emph{Lux Radio Theatre} adaptation (while Moore and McNally do not appear in the film), and that Ritter's involvement was likely not a coincidence due to the \emph{Lux Radio Theatre}'s tendency to ``feature as many of the original stars'' as possible.

To facilitate research on this task, we release \silver{}, a dataset of 34K examples automatically produced from naturally-occurring anchored links that point to specific subsections of a target Wikipedia article.
These examples reflect relevance judgments made by the authors of the source webpage, who are incentivized to help readers by directly linking to relevant sections of target webpages.

Although modeling the distribution of naturally-occurring anchored links may already be useful to readers, constraints on webpage authors bias the linking behavior found in naturally-occurring anchored links. In particular, authors of existing anchored links were constrained by browser conventions to targeting pre-defined anchors in the target webpage (e.g., in Wikipedia, anchored links may only point to subsection headers).
As a result, the existing anchored links in Wikipedia generally refer to a particular set of topics that are broad enough to warrant entire subsections, but too narrow to have a dedicated article. In addition, emerging web standards enable linking to arbitrary spans of text in a target webpage, further motivating the exploration of finer-grained anchors.\footnote{\href{https://wicg.github.io/scroll-to-text-fragment}{wicg.github.io/scroll-to-text-fragment}} Lastly, authors may miss more subtly useful anchors since finding them is a time-consuming process.

To model reader relevance judgments beyond these constraints placed on authors of existing anchored links, we collect \gold{}, an evaluation set of 443 human-constructed examples where annotators manually read through target webpages and selected the paragraph that is most relevant to an unanchored link. Qualitative analysis reveals a variety of differences between links in \silver{} and \gold{}---each dataset reflects different and complementary types of linking behavior, both of which help readers find related information.

We benchmark a variety of baselines built on BM25, finding that relying on lexical overlap alone is insufficient for the task. We also evaluate a T5-based ranking approach and investigate transfer learning from the section-level judgments of \silver{} to the paragraph-level judgments of \gold{}. Overall, our best models achieve 82.84\% accuracy on \silver{} and 31.15\% accuracy on \gold{}. We have released our data at \href{https://nelsonliu.me/papers/wikipedia_anchors}{nelsonliu.me/papers/wikipedia\_anchors}.

\section{Task Definition}

Given a source webpage, a link within the source webpage, and the link's target webpage, the task is to identify the anchor (i.e., part of the target page) that is relevant to the link within the source webpage. See Figure~\ref{fig:example} for an example, where the link ``{\color{examplepurple} Lux Radio Theatre}'' occurs on the source webpage \texttt{Pickup on South Street} and points to the target webpage \texttt{Lux Radio Theatre}.

Formally, we are given an input link $\ell_{s
\rightarrow t}$ that occurs in the source webpage $s$ between character indices $(b_s, e_s)$ and points to the target webpage $t$. Let $c$ be a set of candidate refined link anchors $c = \{c_1, c_2, ..., c_n\}$, where each individual candidate link anchor $c_{i \in n} = (b_t, e_t)$ is a span in the target webpage $t$ with beginning $b_t$ and end $e_t$ character indices. The goal of the task is to select the candidate link anchor $c^{*} \in c$ that is most informative to the reader of the source context of the link $\ell_{s \rightarrow t}$.

The particular choice of the candidate link anchors $c$ is an application-specific design decision. In conventional webpages, the candidate anchors $c$ are the elements within the target webpage with a non-null \texttt{id} attribute. However, we can alternatively define $c$ to be the set of all paragraphs, sentences, or even spans in the target webpage. Note that finer-grained link candidates are not necessarily better---there is a natural trade-off between candidate granularity and the subjectivity of the task. In this work, we focus on paragraph-level anchor candidates, which we qualitatively find to be granular enough to be useful with a manageable level of subjectivity.

Performing this task well challenges models to jointly reason over the lengthy source webpage and the large set of candidates to determine their implicit relations and identify parts of the target webpage that are relevant. 
For example, in Figure~\ref{fig:example}, the unanchored link ``Lux Radio Theatre'' occurs in the source webpage \texttt{Pickup on South Street} and points to the target webpage \texttt{Lux Radio Theatre}.
The link's local context in the source webpage describes that ``Ritter co-starred alongside Terry Moore and Stephen McNally in a Lux Radio Theatre presentation of the story''. Reading other parts of the source webpage reveals that \emph{Pickup on South Street} is a ``1953 Cold War spy film noir'' and that Ritter is a star of the film. Together, this evidence identifies the most relevant candidate anchor as the paragraph describing Lux Radio Theatre's desire to ``feature as many of the original stars of the original stage and film productions as possible'' (in this case, Ritter).

\paragraph{Evaluation.} We evaluate models by their accuracy on the task. Examples often have multiple acceptable anchors (e.g., to account for subjectivity), and we accept a prediction as correct if it matches any of the ground-truth labels.

\section{Dataset Collection and Analysis}

To facilitate research on this task, we construct and release two datasets derived from Wikipedia links (i.e., links in a Wikipedia article that point to other Wikipedia articles). \silver{} contains examples produced from naturally-occurring anchored links on Wikipedia, which point to a specific subsection of the target Wikipedia article. These naturally-occurring links reflect author-driven linking decisions. On the other hand, \gold{} is an evaluation dataset with human-annotated paragraph-level anchors, which reflect reader-driven linking preferences. Dataset statistics are presented in Table~\ref{tab:dataset_statistics}.

\begin{table*}
    \centering
    \footnotesize
    \begin{tabular}{lrrrr}
        \toprule
        & \multicolumn{3}{c}{\silver{}} & \multirow{2}{*}{\gold{}}\\
        \cmidrule(lr){2-4}
        & Train & Dev. & Test\\
        \midrule
        \# Examples & 27,187 & 3,398 & 3,398 & 443 \\
        \# Unique Relevant Anchors & 6.0 $\pm$ 9.0 & 6.0 $\pm$ 9.0 & 5.7 $\pm$ 7.0 & 1.3 $\pm$ 0.5 \\ 
        \# Candidate Anchors & 64.1 $\pm$ 43.1 & 64.1 $\pm$ 42.9 & 63.9 $\pm$ 42.9 & 44.9 $\pm$ 31.5\\
        Relevant Anchor Position & 0.50 $\pm$ 0.24  & 0.51 $\pm$ 0.24 & 0.50 $\pm$ 0.25 & 0.43 $\pm$ 0.27 \\
        \# Source Tokens & 2374 $\pm$ 2872 & 2370 $\pm$ 2855 & 2412 $\pm$ 2994 & 2750 $\pm$ 3141 \\
        Link Position & 0.44 $\pm$ 0.29 & 0.44 $\pm$ 0.29 & 0.44 $\pm$ 0.29 & 0.42 $\pm$ 0.31\\
        \bottomrule
    \end{tabular}
    \caption{\silver{} and \gold{} statistics ($\mu \pm \sigma$). ``Relevant Anchor Position'' and ``Link Position'' denote the mean normalized position of relevant anchors amongst candidate anchors and the link in the source webpage.}
    \label{tab:dataset_statistics}
\end{table*}

Each example in our datasets consists of a Wikipedia link in its source article and a list of candidate anchors with associated relevance judgments. The candidate anchors are the paragraphs of the target article.

\subsection{Identifying Links Worth Refining}

We are most interested in refining links that point to long and multi-faceted target articles; readers have comparatively less to gain from refining links that point to very short or specific articles since simply skimming the entire article is more likely to be quick and effective.

To construct a set of target articles whose in-links are likely to benefit from anchoring,  we filter Wikipedia to keep long and multi-faceted articles. We remove articles with minimal prose, including ``List of'' articles, disambiguation pages, and articles where links comprise more than half of the total text. We also remove articles with fewer than 500 tokens or less than five subsections (excluding trivial subsections such as ``\texttt{References}'' or ``\texttt{See Also}'', which almost never contain useful candidate anchors). In addition, we remove articles that are rarely linked to (less than 25 links across Wikipedia point to this articles), since these articles are often quite niche and primarily serve to define rare entities. Finally, we remove articles that are commonly linked to (more than 5K links across Wikipedia point to this articles), since these articles are often quite broad and have little relation to their linking contexts (e.g., links to articles about countries).

\begin{table*}[!ht]
\centering
\resizebox{\textwidth}{!}{
\begin{tabular}{>{\raggedright}lp{5cm}p{5cm}p{6cm}p{5.5cm}}
\toprule
& \textbf{Source Article} & \textbf{Link in Source Context} & \textbf{Target Article} & \textbf{Anchor Section} \\
\midrule
1. & \texttt{Snorkeling} (British and Commonwealth English spelling: snorkelling) is the practice of swimming on or through a body of water while equipped with \dots & Snorkeling ({\color{examplepurple} \ul{British and Commonwealth English spelling}}: snorkelling) is the practice of swimming on or through a body of water while equipped with \dots & \texttt{American and British English spelling differences}: Despite the various English dialects spoken from country to country and within \dots & \textbf{Doubled consonants} > \textbf{Doubled in British English}: \dots In British English, however, a final -l is often doubled even when the final syllable is unstressed \dots \\
\midrule
2. & The \texttt{Reunion Society of Vermont Officers} was an organization of American Civil War veterans\dots & \textbf{Political influence}: \dots restrict governors to two one-year terms. When two-year terms were introduced, the party limited {\color{examplepurple} \ul{governors}} to a single term\dots & A \texttt{governor} is an administrative leader and head of a polity or political region, ranking under the head of state and in some cases, such as governors-general, as the head of state's official \dots & \textbf{Other modern countries in North America} > \textbf{United States}: In the United States, the title "Governor" refers to the head of each state or insular territory \dots\\
\midrule
3. & The \texttt{Westminster Palace Hotel} was a luxury hotel in London, located in the heart of the political district \dots & \textbf{Opening}: The Hotel opened in 1860 \dots It had all the latest technology, including being the first hotel in London with {\color{examplepurple} \ul{hydraulic lifts}}, advertised as able to \dots & An \texttt{elevator} or lift is a cable-assisted, hydraulic cylinder-assisted, or roller-track assisted machine that vertically transports people or freight between floors, levels, or decks of a building \dots & \textbf{History} > \textbf{Industrial Era}: \dots The hydraulic crane was invented \dots for use at the Tyneside docks for loading cargo. They quickly supplanted the earlier steam-driven elevators \dots \\
\bottomrule
\end{tabular}
}
\caption{Examples from \silver{} require diverse types of reasoning.
The first example requires understanding that the British spelling ``snorkelling'' differs from the American spelling \texttt{snorkeling} because of the repeated consonant ``ll'', and that the link exists to provide further details about spelling variations due to double consonants.
The second example requires understanding that the source context is (implicitly) about governors in the United States and localizing this information in the target article.
The third example requires inferring that the opening of the \texttt{Westminster Palace Hotel} is related to other events in the history of the \texttt{elevator} (the anchor section).
{\color{examplepurple} \ul{Link text}} is colored and underlined, \texttt{article titles} appear in monospace font, and \textbf{section headings} are bolded, with > used to denote nesting.
}
\label{tab:silver_examples}
\end{table*}

\subsection{Collecting \silver{}}

We use naturally-occurring anchored links from English Wikipedia (July 20, 2022 snapshot) for training and evaluation data. Since naturally-occurring anchored Wikipedia links target a particular \emph{section} within the target article, we convert these section-level annotations to paragraph-level annotations by considering all paragraphs in the linked section to be acceptable refined anchors.

\paragraph{Step 1: Collect anchored links.} We extract the anchored links that point to valid target articles ($\sim$160K total links).

\paragraph{Step 2: Deduplicate.} Anchored links in Wikipedia are often duplicated across pages. For example, many articles about biblical manuscripts (e.g., \texttt{Papyrus 44}) will contain a link with text ``Gregory-Aland'', e.g., ``Papyrus 44 (in Gregory-Aland numbering) ... is an early copy of the New Testament in Greek''. These links almost always point to the target Wikipedia article \texttt{Biblical Manuscript}, specifically the subsection ``Gregory–Aland'', which describes a particular system for cataloging biblical manuscripts.
When two anchored links in our set of 160K have the same link text and point to the same target article, they also point to the same section of the target article over 96\% of the time. Consequently, we deduplicate the data such that no two links have the same link text and point to the same target article, resulting in 39,312 deduplicated examples.

\paragraph{Step 3: Filter trivial examples.} Wikipedia maintains editorial guidelines that determine whether topics are presented on a dedicated standalone page or within a larger page about a broader topic. Anchored links in Wikipedia are commonly used to point readers directly to topics that do not have a standalone page (e.g., Gregory-Aland numbering in the prior example). We find that such links are often trivial to refine---when an anchored link's text matches the heading of a subsection in the target article, the link points to this matching subsection 94\% of the time. As a result, we remove examples if the link text matches the linked subsection's heading.

\silver{} has 33,983 total examples. We use 80\% of this data for training, 10\% for validation, and 10\% for testing. Table~\ref{tab:silver_examples} presents examples from the dataset; successful anchor prediction requires diverse types of reasoning.

\subsection{Collecting \gold{}}

\begin{table*}[!ht]
\centering
\resizebox{\textwidth}{!}{
\begin{tabular}{>{\raggedright}lp{5cm}p{5cm}p{6cm}p{7cm}}
\toprule
& \textbf{Source Article} & \textbf{Link in Source Context} & \textbf{Target Article} & \textbf{Relevant Anchor Paragraph} \\
\midrule
1. & \texttt{History of the New York Giants (1994–present)}: The New York Giants, an American football team which currently plays in the NFL's National Football Conference, have qualified for the postseason seven times since 1994. With the retirement of Phil Simms and Lawrence \dots & \textbf{Tom Coughlin era: 2004–2015} > \textbf{2004 NFL Draft and arrival of Eli Manning}: Accorsi \dots saw University of Mississippi quarterback Eli Manning as a similar talent. Manning's brother Peyton, and his father {\color{examplepurple} \ul{Archie}}, had already established successful careers as NFL quarterbacks \dots & \texttt{Archie Manning}: Elisha Archibald Manning III (born May 19, 1949) is a former American football quarterback who played in the National Football League (NFL) for 13 seasons, primarily with the New Orleans Saints. He played for the Saints from 1971 to 1982 and also had brief stints with the Houston Oilers and Minnesota Vikings \dots & \textbf{NFL Career}: Manning was selected to the Pro Bowl in 1978 and 1979 \dots He ended his 13-year career having completed 2,011 of 3,642 passes for 23,911 yards, 125 touchdowns, and 173 interceptions \dots His 2,011 completions ranked 17th in NFL history upon his retirement \dots \\
\midrule
2. & \texttt{Jangle} or jingle-jangle is a sound typically characterized by undistorted, treble-heavy electric guitars (particularly 12-strings) played in a droning chordal style (by strumming or arpeggiating). The sound is mainly associated with pop music~\dots & \textbf{Popularization}: \dots~the Beatles~\dots~are commonly credited with launching the popularity of jangle pop~\dots~the Beatles inspired many artists to purchase Rickenbacker 12-string guitars through songs such as \dots~{\color{examplepurple} \ul{``Ticket to Ride''}}~\dots & \texttt{Ticket to Ride (song)}: ``Ticket to Ride'' is a song by the English rock band the Beatles, written primarily by John Lennon and credited to Lennon–McCartney. Issued as a single in April 1965, it became the Beatles' seventh consecutive number 1 hit in the United Kingdom and their third~\dots & \textbf{Recording}: The song's main guitar riff was played by Harrison on his Rickenbacker 12-string guitar~\dots~According to Harrison, however, the Rickenbacker riff was his own idea, based on the way Lennon strummed the chord when introducing the song to the band~\dots \\
\midrule
3. & \texttt{John Quinn (collector)}: John Quinn~\dots~was an Irish-American cognoscente of the art world and a lawyer in New York City who fought to overturn censorship laws restricting modern literature and art from entering the United States.~\dots & \textbf{Biography}: In the early 1920s Quinn represented Margaret Anderson and Jane Heap for their publication in The \emph{Little Review} of serial portions of {\color{examplepurple} \ul{James Joyce's}} \emph{Ulysses}, which the U.S. Post Office had found ``obscene''~\dots & \texttt{James Joyce}: James Augustine Aloysius Joyce (2 February 1882 – 13 January 1941) was an Irish novelist, short story writer, poet, and literary critic. He contributed to the modernist avant-garde movement and is regarded as one of the most influential and important writers of the 20th century~\dots & \textbf{1920–1941: Paris and Zürich} > \textbf{Paris} > \textbf{Publication of Ulysses}: \dots~With financial backing from the lawyer John Quinn, Margaret Anderson and~\dots~Jane Heap had begun serially publishing \dots suppressed as obscene and potentially subversive \dots trial proceedings continued until February 1921, when both Anderson and Healy, defended by Quinn~\dots \\
\bottomrule
\end{tabular}
}
\caption{Anchors in \gold{} play diverse roles in their source context.
The first example requires understanding that \texttt{Archie Manning}'s NFL career accomplishments (the annotated anchor paragraph) is the particular aspect of the target page that is compared in this context (namely, the hope that Eli could replicate his father's success). In contrast, other relations between Eli and Archie (e.g., father-son), are irrelevant in this setting.
and the prospect of Eli replicating his success, is the aspect of the target page that is salient to link's particular source context (rather than relations like ``father'').
In the second example, models must recognize that ``Rickenbacker 12-string guitars'' connect the source article \texttt{Jangle} with the target article \texttt{Ticket to Ride}.
The third example requires identifying coreferent events; the link's source context discusses \texttt{John Quinn}'s role in the trial proceedings around James Joyce's \emph{Ulysses}, and the annotated relevant anchor paragraph further expounds on this event.
{\color{examplepurple} \ul{Link text}} is colored and underlined, \texttt{article titles} appear in monospace font, and \textbf{section headings} are bolded, with > used to denote nesting.
}
\label{tab:gold_examples}
\end{table*}

The limitations of conventional Internet links greatly influence the distribution of naturally-occurring anchored links in \silver{}, since conventional Internet links are constrained to either point to (i)~the whole target webpage or (ii)~an anchor within the target webpage (section headings, in the case of Wikipedia). As a result, existing anchored links in Wikipedia tend to refer to topics that are broad enough to warrant entire sections.
Similarly, authors may want to link to a particular portion of the target webpage, but there is no suitable anchor within the target webpage.
As a result, they resort to creating an unanchored link to an entire target webpage. Authors may also miss more subtly useful anchors that are time-consuming to find.

We collect \gold{} to model reader relevance judgments beyond the distribution of naturally-occurring anchored links. Given links in their source article context, annotators are asked to select the target article paragraph (i.e., the candidate anchor) that is most informative to the link's particular source context.

Human relevance judgments are inherently subjective---readers have varying levels of prior knowledge about the source and target articles, which affects which parts of a target article they might find informative. For example, readers with minimal background often derive the most value from reading the first paragraph of the target page, since this \emph{lead paragraph} is designed to provide broad overview and definition of the subject. On the other hand, readers with more prior knowledge about the subject may find the lead paragraph redundant. To control for these disparities in prior knowledge, we show annotators the target article's lead paragraph so they at least have a shared lower-bound of knowledge about the subject.

In many cases, there does not exist an anchor that is more useful to the reader than the lead paragraph, and these links are best left unanchored. For example, many Wikipedia links exist solely to define the linked term, and there is no other relation between the source and target articles. Inferring when this is the case is a surprisingly difficult problem on its own, and it is difficult to filter out unanchorable links. In fact, the vast majority (more than 98\%) of links were considered unanchorable by annotators, leading to a low yield of anchor annotations.
We believe that this low number is specific to Wikipedia due to its encyclopedic nature and would be much higher in the open web. Due to the low annotation yield, we only use \gold{} as an evaluation set. We also find the decision of anchorable vs. unanchorable much more subjective than the choice of the anchor, so we leave this aspect to future work; every example in our data has at least one acceptable anchor.

Table~\ref{tab:gold_examples} presents examples from \gold{}---successful anchor prediction often requires reasoning about what aspects of entities are shared and compared in a given context.

\paragraph{Annotation process.} For each example, annotators were first shown the source article, the link in its source context, and the first paragraph of the target article. Annotators were asked to read as much of the source article as necessary in order to understand the link and in its context. The first paragraph of the target article was shown to give annotators a shared baseline of basic information about the subject of the target article.

Equipped with an understanding of the link in its source context and some baseline knowledge about the subject of the target article, annotators were then asked to read through the target article and select the non-lead paragraph that is most informative to the source linking context, if such a paragraph exists.

Given that many links are best left unanchored, we use a two-stage annotation approach to control annotation costs. In the first stage, we elicited a single annotation for $\sim$26K unanchored links, resulting in 443 examples with anchors. In the second stage, we collected two more annotations for each of the 443 examples produced from the first stage; collecting multiple annotations per example is crucial for accounting for the inherent subjectivity of relevance judgments.

Annotation was performed by the authors of this paper and a team of 18 hired annotators.
Annotators participated in a 30-minute training session and received regular feedback from the authors throughout several annotation pilot studies.
This annotation task requires significant time and attention, since annotators must read enough of the source article to understand the link in its context before reading through each target article paragraph and selecting the one that is most related. On average, annotation took 4.45 minutes per link (approximately 2K hours in total).

\paragraph{Annotation quality.}
To assess the annotation quality and estimate the amount of label noise, we manually review a sample of 100 annotations. For each annotation, we judge whether a reader would conceivably find the annotated anchor relevant to the link's original source context. We find 92\% of our sampled annotations to be of acceptable quality.

\begin{figure}
  \centering
  \includegraphics[width=0.9\columnwidth]{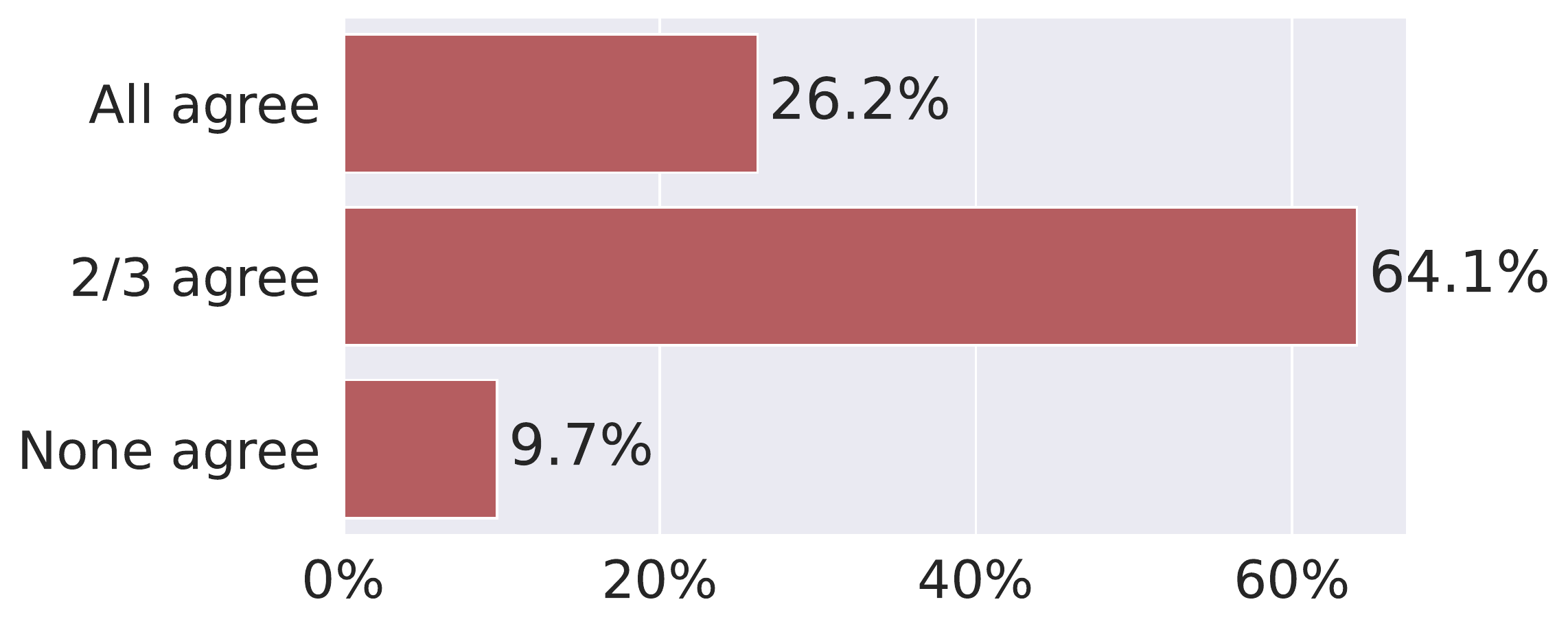}
  \caption{Agreement statistics on \gold{}. Despite the inherent subjectivity of the task, readers often make similar anchoring decisions.}
  \label{fig:agreement}
\end{figure}

\paragraph{Subjectivity analysis.}
We expect reader-driven relevance judgments to be subjective, since readers have varying levels of background knowledge. To quantify this subjectivity, we measure agreement statistics of annotator decisions
(i.e., which candidate anchor is most relevant, or that none exists; Figure~\ref{fig:agreement}).
We find that readers often agree on a small set of potential link candidates. Much like in other subjective NLP tasks such as text generation and information retrieval, while the \emph{best} prediction is subjective, there is little ambiguity in poor predictions.
Our evaluation metrics also account for subjectivity by accepting predictions that match any of the provided relevance judgments.
\gold{} Cohen's Kappa is $\kappa$ = 0.47, indicating moderate agreement.

\paragraph{\silver{} and \gold{} exhibit different linking behavior.} 
We analyze \silver{} and \gold{} links to compare their different linking behaviors. We find that links in \gold{} target articles about named entities, particularly people and organizations, at a much higher rate than \silver{} links. This occurs because these entities generally have standalone Wikipedia articles, and naturally-occurring anchored links can only target subsections of existing Wikipedia pages. Consequently, the named entities that \emph{do} occur in \silver{} are generally subclasses of a broader entity (e.g., the link  ``1977 Tokyo Motor Show'' points to the subsection ``1977'' of the article \texttt{Tokyo Motor Show}).

In addition, these differences in the target webpage distribution affect the type of information that is found to be relevant in each dataset. Since links in \gold{} point to broader topics (given that they are notable enough to have a standalone page), annotators tend to focus on higher-level semantic relations between the subjects of the source and target article (e.g., the relation between \texttt{Pickup on South Street} and \texttt{Lux Radio Theatre} in Figure~\ref{fig:example}). In contrast, links in \silver{} often connect their \emph{local} context with the target article. For example, some \silver{} links help users navigate lengthy target articles (e.g., ``Europa League quarter-final second leg'' pointing to the section ``quarter-final'' in the target article \texttt{2012–13 UEFA Europa League}). Other \silver{} links express more-abstract relations between the link text and the target article (e.g., ``made to write with his right hand'' linking to the section ``Negative connotations and discrimination [of left-handed people]'' in the target article \texttt{Handedness}).

\section{Baseline Methods}

\paragraph{BM25 baselines.} Our first set of baselines is designed to measure whether lexical overlap between the source webpage and candidate anchors are sufficient for high performance on the task. These baselines simply predict the candidate anchor that has the highest BM25 score with either the link's local context in the source webpage (BM25-Context) or the source webpage title (BM25-Title).

When all candidate anchors share zero words with either the webpage title or the local link context, we output the candidate corresponding to the lead paragraph of the target webpage, which always at least provides basic information about the linked term.

\paragraph{RankT5.} T5 is a pre-trained sequence-to-sequence model trained on a large web corpus \citep{JMLR:v21:20-074}. To predict anchors with T5, we frame the task as a paragraph ranking problem, following the RankT5 procedure of \citep{zhuang2022rankt5}. More formally, the input to the ranking task is a query $q$ and a set of $n$ candidate documents $D = (d_1, ..., d_n)$. We are also given binary relevance labels $\mathbf{y} = (y_1, ..., y_n)$ (where $y_{i} \in \{0, 1\}$) for each of the $n$ candidate documents in $D$. A ranking model $f$ takes a query-document pair as input and produces a real-valued ranking score $\hat{y_i} = f(q, d_i) \in \mathbb{R}$. The documents in $D$ are sorted by their ranking scores with respect to a query $q$, and models are trained to optimize ranking losses to assign higher ranking scores to relevant documents and lower scores to non-relevant documents.

Each example in our dataset contains a link and its source webpage (the query $q$), a set of candidate anchors (the documents $D$ to be ranked), and labels indicating the acceptable candidates (the relevance labels $\mathbf{y}$). The input to T5 is a string containing information about the query-document pair to be ranked. More specifically, the input  contains: (i)~the source and target webpage titles, (ii)~excerpts of the source and target webpage lead paragraphs, (iii)~the text surrounding the link in the source webpage, (iv)~the heading of the section containing the link, (v)~the text of the candidate anchor, and (vi)~the heading of the section containing the candidate anchor.

Following \citet{zhuang2022rankt5}, we use the unnormalized decoder logits of a special unused token in the T5 vocabulary (``\texttt{<extra\_id\_10>}'') as the ranking score.
During fine-tuning, each dataset example is split into $k$ lists, where $k$ is the number of anchors labeled as relevant. Each list has $m$ elements and contains a single acceptable anchor and $(m-1)$ randomly-sampled unacceptable candidates. Due to hardware memory constraints, we set the list size $m$ to be 36 and fine-tune on batches of 32 lists (32 $\times$ 36 = 1152 total sequences per batch).
The model is optimized with a listwise softmax cross-entropy loss \citep{10.1145/3341981.3344221}, which jointly considers the ranking scores of all documents in a list.
During inference, we compute the ranking score for all candidate anchors and output the candidate with the the highest score. See Appendix~\ref{sec:rankt5_implementation_details} for further implementation details.

\section{Results and Analysis}

Table~\ref{tab:results} presents baseline results. The majority label baseline predicts the candidate anchor index that is most frequently marked as relevant in the training data, and the random label baseline predicts a random candidate anchor. Majority and random label baselines achieve higher performance on \silver{} than \gold{} because, on average, examples in \silver{} have more unique relevant anchors (Table~\ref{tab:dataset_statistics}).

\paragraph{\silver{}.} The first two columns of Table~\ref{tab:results} present the performance of all baseline methods on \silver{}. The BM25 baselines perform quite poorly, with BM25-Title only slightly outperforming the random baseline. Lexical cues from the link's local source context are more predictive than using the source webpage title, supporting qualitative observations that links in \silver{} are often used to define terms in a particular source context, rather than expounding upon higher-level relationships between the subjects of the source and target webpages.

The T5-based ranking models perform much better than the BM25 baselines or the trivial baselines. Increasing the size of the pre-trained model modestly improves performance.

\paragraph{\gold{}.}
The third column of Table~\ref{tab:results} presents baseline results on \gold{}. The BM25 baselines achieve higher performance on \gold{} than \silver{}, suggesting that lexical overlap has higher correlation with reader-driven relevance judgments than author-driven relevance judgments. We also find that BM25-Title outperforms BM25-Context on \gold{}, potentially because reader-driven relevance judgments tend to focus on higher-level connections between the source and target webpages, rather than lower-level relations that are specific to a particular local source context. The different performance trends of the BM25 baselines between \silver{} and \gold{} provide further evidence that these two datasets exhibit different linking behavior, though both are valuable for localizing different types of information in different types of target pages.

\begin{table}
\setlength{\tabcolsep}{8pt}
  \centering
  \begin{tabular}{lccc}
    \toprule
    & \multicolumn{2}{c}{\makecell{\textsc{Author}\\\textsc{Anchors}}} &  \makecell{\textsc{Reader}\\\textsc{Anchors}}\\
    \cmidrule(lr){2-3} \cmidrule(lr){4-4}
    & Dev & Test & Test  \\
    \midrule
    BM25-Title & 11.5 & 12.0  & 27.1 \\
    BM25-Context & 19.4 & 18.8 & 22.6 \\
    RankT5-Small & 74.2 & 74.6  & 29.8 \\
    RankT5-Base & 80.0 & 80.6 & 31.2 \\
    RankT5-Large & 82.2 & 82.8  & 30.5 \\
    \midrule
    Majority Label & 12.7 & 14.0  & ~~7.7 \\
    Random Label & 11.5 & 11.3  & ~~4.6 \\
    \bottomrule
\end{tabular}
  \caption{Accuracy of various baselines on \silver{} and \gold{}. Note that the RankT5 models are trained on \silver{}.}
  \label{tab:results}
\end{table}

\paragraph{Transfer learning from \silver{} to \gold{}.} To establish baseline performance of RankT5 models on \gold{}, we train them on \silver{} and evaluate them on \gold{}; these results are also found in Table~\ref{tab:results}. RankT5 models trained on \silver{} transfer poorly to \gold{} due to the differences in the link distributions. For example, section headings are often quite informative for anchoring links in \silver{} since these links are originally section-level. However, we find that models are overreliant on heading text when evaluated on \gold{}, which contains paragraph-level relevance judgments. Models often prefer to select a paragraph within a section whose heading is related to the link's source context, even when the paragraph content is uninformative.

\section{Related Work}
\paragraph{Ranking tasks.}
The form of our task definition resembles well-established ranking tasks. Unsurprisingly, we chose baselines that were designed for such tasks such as BM25 \citep{robertson-1994-okapi} and RankT5~\citep{zhuang2022rankt5}. Despite the similarities in form, anchor prediction involves very different kinds of inferences, since source context is much longer and the intent is more subtle than a traditional `query' in information retrieval.

\paragraph{Anchoring other linking tasks.}

We use Wikipedia links in this work to serve as an initial instantiation of the anchor prediction task. However, the notion of anchoring a link is applicable to any type of ``link'' between a source context and a target document. Examples of other types of links that could benefit from anchors include (i)~information retrieval~\cite{manning08retrieval} which links queries to documents, (ii)~citation recommendation~\cite{McNee2002OnTR} which link scientific papers to other scientific papers, and (iii)~retrieval-augmented models~\cite{Guu2020REALMRL} which link model inputs to useful documents. In all such cases, the consumer of such links (either users or models) could benefit from a more precise destination that is anchored to parts of the document that are more useful than lead paragraph.

The CL-SciSumm shared task on scientific document summarization is the closest prior line of work \citep{chandrasekaran-etal-2020-overview-insights}. In subtask 1A, systems are given a citation within the context of a scientific paper and must identify the spans of text in the cited research paper that most accurately reflect the original citing context. The annotated spans are either sentence fragments, a single full sentence, or several consecutive sentences.

The input-output format of subtask 1A of the CL-SciSumm shared task is similar to anchor prediction, but the tasks differ in their practical utility and research  goals. In particular, the annotations in subtask 1A of the CL-SciSumm shared task are not designed for use as a standalone application, but rather as input for a downstream summarization system. In contrast, the goal of the anchor prediction task is to help readers effectively find information on the web.
Furthermore, relations between citing and cited papers are often quite narrow---\citet{cohan-etal-2019-structural} argue that almost all citations can be classified into 3 categories: ``Background Information'', ``Method'', and ``Result Comparison''. On the other hand, our data reflects the comparatively richer linking behavior found on the Internet and contains greater diversity of relations between linking and linked documents.

\paragraph{Improving content-based link prediction.}
There exist content-based approaches to predicting all of the types of previously-aforementioned links~\cite{Bhagavatula2018ContentBasedCR, Logeswaran2019ZeroShotEL}. These approaches typically use the leading text as the heuristic content to be encoded by the model. We posit that modeling and using anchors instead to represent the content would lead to more accurate link prediction.

\section{Conclusion and Future Work}

In this work, we introduce the task of anchor prediction, where the goal is to refine the target of an  input link by selecting the candidate anchor that is most related to the source linking context. To facilitate research on this task, we introduce \silver{}, which contains 34K examples derived from existing anchored links in Wikipedia. These naturally-occurring anchored links reflect relevance judgments by webpage authors, who are incentivized to create useful and informative links. To complement these author-driven relevance judgments, we collect \gold{}, an evaluation set with human-annotated relevance judgments from readers.
Although our analysis reveals that examples in \silver{} and \gold{} exhibit very different linking behavior, they are both useful to readers. We benchmark a variety of baselines on \silver{} and investigate transfer learning from \silver{} to \gold{}, finding significant room for improvement.

This work provides an initial instantiation of the anchor prediction task, and we expect a variety of interesting opportunities and challenges to arise from extending this task to other types of links. For example, anchoring citations in scientific or legal documents requires expert annotators, and anchoring links on the open web may require greater flexibility than paragraph-level anchor candidates.

\section*{Acknowledgements}
We are grateful to the team of 18 annotators who participated in data collection---this work
would not have been possible without them. We also thank  Muqthar Mohammad and Kiranmai Chennuru for coordinating the annotation process. Finally, we thank Ming-Wei Chang, Chaitanya Malaviya, Pete Shaw, and Grant Wang for feedback and discussions that helped improve this work.

\section*{Limitations}
A primary goal of this work is to introduce the anchor prediction task and construct initial benchmarks for investigation. As a result, \silver{} and \gold{} make a variety of compromises to the external validity of the task.
For example, we use Wikipedia links in this work as an initial proof-of-concept for this task, but the encyclopedic nature of Wikipedia likely differs significantly from links found in the open web.
In addition, we specifically filter to find links that point to long, multi-faceted webpages because (i)~we believe these examples pose interesting language understanding challenges and (ii)~readers stand to gain the most from refined anchors to these pages.
Lastly, because many links are best left unanchored, any practically-useful anchor prediction system must learn to abstain from prediction; we leave this aspect for future work.

Since our focus is laying the groundwork for future research on anchor prediction, we use English Wikipedia. Despite our focus on English, the data collection pipeline used to produce \silver{} can be readily applied to other languages. We leave exploration of multilingual and cross-lingual anchor prediction for future work.

Lastly, our models consider a fixed candidate anchor granularity (i.e., paragraphs). Modeling the level of granularity for a particular task or context is a promising direction for future work.

\bibliography{custom}
\bibliographystyle{acl_natbib}

\appendix

\section{RankT5 Implementation Details}
\label{sec:rankt5_implementation_details}

Our RankT5 models are fine-tuned for 100,000 steps with the Adafactor optimizer \citep{pmlr-v80-shazeer18a} with a constant learning rate of 0.001 and a dropout rate of 0.1. The model is optimized with a listwise softmax cross-entropy loss \citep{10.1145/3341981.3344221}:

$$
  L(\mathbf{y, \hat{y}}) = -\sum_{i=1}^{m} y_i \log \left( \frac{e^{\hat{y}_i}}{\sum_{i'} e^{\hat{y}_i'}} \right)
$$

Training was performed on a cluster 64 Google Cloud TPUs (v3). Our models are implemented in the JAX-based T5X library \citep{roberts2022t5x}.

\end{document}